\documentclass{sig-alternate}

\usepackage[
  pass,
]{geometry}

\usepackage{bbm}
\usepackage{hyperref}
\usepackage{caption}
\usepackage{color}
\usepackage{xspace}

\DeclareCaptionType{copyrightbox}
\usepackage{subcaption}
\usepackage{booktabs}

\newcommand{\xhdrNoPeriod}[1]{\vspace{1mm}\noindent{{\bf #1}}}

\newcommand{\sectionrule}{\addlinespace[0.5ex]}

\begin{document}

\numberofauthors{1} 
%
\newcommand{\authspace}{\hspace*{.2in}}
\author{
Subhabrata Mukherjee$^\dag$ \authspace  Gerhard Weikum$^\dag$  \authspace \mbox{Cristian Danescu-Niculescu-Mizil$^\ddag$}\\
 \affaddr{$^\dag$Max Planck Institute for Informatics, $^\ddag$Max Planck Institute for Software Systems} \\
 \email{smukherjee@mpi-inf.mpg.de, weikum@mpi-inf.mpg.de}, cristian@mpi-sws.org}

\title{People on Drugs: \\ Credibility of User Statements in Health Communities}

\newcommand{\squishlist}{
   \begin{list}{$\bullet$}
    { \setlength{\itemsep}{0pt}      \setlength{\parsep}{3pt}
      \setlength{\topsep}{3pt}       \setlength{\partopsep}{0pt}
      \setlength{\leftmargin}{1.5em} \setlength{\labelwidth}{1em}
      \setlength{\labelsep}{0.5em} } }
\newcommand{\squishlisttwo}{
   \begin{list}{$\bullet$}
    { \setlength{\itemsep}{0pt}    \setlength{\parsep}{0pt}
      \setlength{\topsep}{0pt}     \setlength{\partopsep}{0pt}
      \setlength{\leftmargin}{0.5em} \setlength{\labelwidth}{0.5em}
      \setlength{\labelsep}{0.5em} } }

\newcommand{\squishend}{
    \end{list} 
}

\newcommand{\comment}[1]

\maketitle

\section*{Abstract}

Online health communities are a valuable source of information for patients and physicians. However, such 
user-generated resources
are often plagued by inaccuracies and misinformation. 
In this work we propose a method for automatically establishing the credibility of user-generated medical statements and the trustworthiness of their authors by exploiting linguistic cues and distant supervision from expert sources.
To this end we introduce a probabilistic graphical model that jointly learns 
user trustworthiness, statement credibility, and language objectivity.
We apply this methodology to the task of extracting rare or unknown side-effects of medical drugs --- this being one of the problems 
where large scale non-expert data has the potential to complement expert medical knowledge.
We show that our method can reliably extract side-effects and filter out false statements, while identifying trustworthy users that are likely to contribute valuable medical information.

\vspace*{-1mm}
\ -\ \textit{Information Filtering}
\ -\ \textit{Natural Language Processing}
\vspace*{-1.5mm}
\terms{Design, Algorithms, Measurement, Experimentation}
\vspace*{-1.5mm}
\vspace*{-0.5mm}

\section{Introduction}

Online social media includes a wealth of topic-specific communities and discussion forums about politics, music, health, and many other domains. User-generated content in such communities offer a great potential for distilling and analyzing facts and opinions.
In particular, online health communities constitute an important source of information for patients and doctors alike,
with 59\% of the adult U.~S. population consulting online health resources \cite{Fox:PewInternetAndAmericanLifeProject:2013},
and nearly half of U.~S. physicians relying on online resources \cite{IMS2014} for professional use.

One of the major hurdles preventing the full exploitation of information from online health communities is the widespread concern regarding the quality and credibility of user-generated content \cite{Peterson:JMedInternetRes:2003,cline2001consumer,white2014health}. To address this issue, this work proposes a model that can automatically assess the credibility of medical statements made by users of online health communities.
In particular, we focus on extracting rare or unknown side-effects of drugs --- this being one of the problems where large scale non-expert data has the potential to complement expert medical knowledge \cite{White2014}, but where misinformation can have hazardous consequences.

The main intuition behind the proposed model is that there is an important interaction between the credibility of a statement, the trustworthiness of the user making that statement and the language used in the post containing that statement. Therefore, we consider the mutual interaction between the following factors:
\squishlist
\item {\em Users:} the overall {\em trustworthiness} (or authority) of a user, corresponding to her status and engagement in the community.
\item {\em Language:} the {\em objectivity}, rationality (as opposed to emotionality), and general quality of the language in the users' posts.
Objectivity is the quality of the post to be free from preference, emotion, bias and prejudice of the author.
\item {\em Statements:} the {\em credibility} (or truthfulness) of medical statements contained within the posts. Identifying accurate side-effect statements is a goal of the model.
\squishend

These factors have a strong influence on each other. Intuitively, a statement is more credible if it is posted by a trustworthy user and expressed using confident and objective language. As an example, consider the following review about the drug Depo-Provera by a senior member of {\small\tt \href{http://www.healthboards.com/}{healthboards.com}}, one of the largest online health communities:\\
``{\small \dots Depo is very dangerous as a birth control and has too many long term side-effects like reducing bone density \dots}''\\
This post contains a credible statement that a side-effect of Depo-Provera is to reduce bone density. Conversely, high subjectivity and emotionality in a post suggest lower credibility of statements and lower believability in that user's contents. A negative example along these lines is:\\
``{\small I have been on the same cocktail of meds (10 mgs. Elavil at bedtime/60-90 mgs. of Oxycodone during the day/1/1/2 mgs. Xanax a day....once in a while I have really bad hallucination type dreams. I can actually ``feel" someone pulling me of the bed and throwing me around. I know this sounds crazy but at the time it fels somewhat demonic.}''\\
Although this post suggests that taking Xanax can lead to hallucination,
the style in which it is written renders the credibility of this statement doubtful.
These examples support the intuition that to identify credible medical statements, we also need to assess the trustworthiness of users and the objectivity of their language. In this work we leverage this intuition through a {\em joint analysis of statements, users, and language}
in online health communities. 

Although information extraction (IE) methods using probabilistic graphical models \cite{Sarawagi2008,KollerFriedman2009} can be used to extract statements from the user posts, they do not account for the inherent bias, subjectivity and misinformation prevalent in the health forums. Furthermore, standard IE techniques \cite{Krishnamurthy2009,Bohannon2012,SuchanekWeikum2013} do not consider the role of language in extracting credible statements. In this work, we perform linguistic analyses to extract stylistic and affective features from user posts. Discourse features help to identify authoritative user statements by examining the usage of modals, inferential conjunctions, hypothetical statements etc; whereas affective features help to identify objective statements by analyzing user emotions in the post like anxiety, depression, esteem, confidence.

The main technical contribution of this paper is a probabilistic graphical model \cite{KollerFriedman2009} which is tailored to the problem setting as to facilitate joint inference over users, language, and statements.
We devise a Markov Random Field (MRF) with individual users, posts, and statements as nodes, as summarized in Figure \ref{fig:triangle}.
The quality of these nodes -- trustworthiness, objectivity, and credibility --
is modeled as binary random variables.
Our method is semi-supervised with a subset of training side-effect statements,
derived from expert medical databases, labeled as true or false.
We use linguistic and user features
that can be directly observed in online communities.
Inference and parameter estimation is done via an EM (Expectation-Maximization)
framework, where MCMC sampling is used
in the \textit{E-step} for estimating the label of unknown statements and in the \textit{M-step} feature weights are computed by Trust Region Newton method~\cite{Lin2008}.

We apply our method to
$2.8$ 
million posts contributed by $15,000$ users of one of the largest online health community {\small\tt \href{http://www.healthboards.com/}{healthboards.com}}. Our model achieves an overall accuracy of $82\%$ in
identifying drug side-effects,
with an improvement of $13\%$ over an SVM baseline using the same features and an improvement of 4\% over a stronger SVM classifier which uses distant supervision to account for feature sparsity.  We further evaluate how the proposed model performs in two realistic use cases: discovering rare side-effects of drugs and identifying trustworthy users in a community.

To summarize, this paper brings the following main contributions:
\squishlist
\item {\em Model:} It proposes a model that captures the interactions between user trustworthiness, language objectivity, and statement credibility
in social media (Section~\ref{sec:model}), and devises a comprehensive feature set to this end (Section~\ref{sec:features});
\item {\em Method:} It introduces a method for joint inference over users, language, and statements (Section~\ref{sec:inference}) by a judiciously designed probabilistic graphical model;
\item {\em Application:} It applies this methodology to the problem of extracting side-effects of medical drugs from online health forums (Section~\ref{sec:experiments});
\item {\em Use-cases:} It evaluates the performance of the model in the context of two realistic practical tasks (Section~\ref{sec:usecases});
\squishend

\section{Overview of the Model}
\label{sec:model}

Our approach leverages the intuition that there is an important interaction between statement credibility,
linguistic objectivity, and user trustworthiness.  We therefore model these factors jointly through a probabilistic graphical model, more specifically a Markov Random Field (MRF), where
each statement, post and user is associated with a binary
random variable. Figure \ref{fig:triangle} provides an overview of our model.  For a given statement, the corresponding variable should have value 1 if the
statement is credible, and 0 otherwise. Likewise, the values of
post and user variables reflect the objectivity and trustworthiness
of posts and users.

\xhdrNoPeriod{Nodes, Features and Labels}
Nodes associated with users and posts have observable features, which can be extracted from the online community.
For users, we derive engagement features (number of questions and answers posted),
interaction features (e.g., replies, giving thanks),
and demographic information (e.g., age, gender).
For posts, we extract linguistic features in the form of discourse markers
and affective phrases.  Our features are presented in details in Section~\ref{sec:features}.
While for statements there are no observable features, we can derive
distant training labels for a subset of statements from expert databases, like the Mayo Clinic,\footnote{\small\tt \href{http://www.mayoclinic.org/drugs-supplements/}{mayoclinic.org/drugs-supplements/}}
which list typical as well as rare side-effects of widely used drugs.

\xhdrNoPeriod{Edges}
The primary goal of the proposed system is to retrieve the credibility label of unobserved statements given \textit{some} expert labeled statements and the observed features by leveraging the mutual influence between the model's variables.
To this end, the corresponding (undirected) edges in the MRF are as follows:
\squishlist
\item each user  is connected to all her posts;
\item each statement is connected to all posts from
which it can be extracted (by state of the art information extraction methods);
\item each user is connected to statements
that appear in at least one of her posts;
\squishend
Configured this way, the model has the capacity to capture important interactions between statements, posts, and users --- for example, credible statements can boost a user's trustworthiness, whereas some false statements may bring it down.  Furthermore, since the inference (detailed in Section \ref{sec:inference}) is centered around the cliques in the graph (factors) and multiple cliques can share nodes, more complex ``cross-talk'' is also captured.  For instance, when several highly trustworthy users agree on a statement and one user disagrees, this reduces the trustworthiness of the disagreeing user.

In addition to establishing the credibility of statements, the proposed system also computes individual likelihoods as a by-product of the inference process, and therefore can output rankings for all statements, users, and posts, in descending order of credibility, trustworthiness, and objectivity.

\begin{figure}
\includegraphics[scale=0.36]{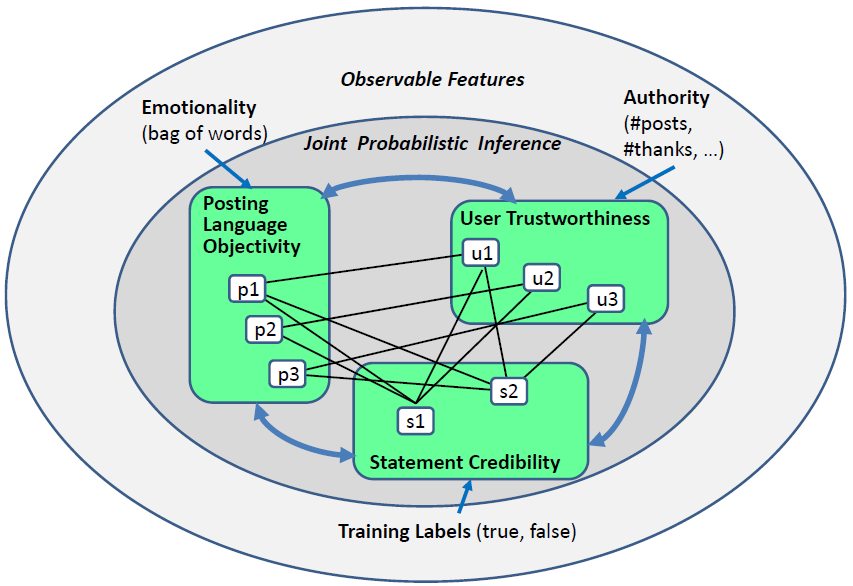}
\caption{Overview of the proposed model, which captures the interactions between statement credibility, post objectivity, and user trustworthiness.}
\label{fig:triangle}
\vspace*{-4mm}
\end{figure}

\section{Features}
\label{sec:features}

\subsection{Linguistic Features}

The language characteristics of a post can convey the author's attitude towards the statements as well as her certainty, or lack thereof~\cite{Coates1987}.
In our model 
we use \textit{stylistic} features
and \textit{affective} features for assessing a post's objectivity and quality.

\xhdrNoPeriod{Stylistic Features} Consider the following user post: \\
{\small ``I heard Xanax \underline{can} have pretty bad side-effects. You \underline{may} have peeling of skin, and apparently \underline{some} friend of mine told me you \underline{can} develop ulcers in the lips also. \underline{If} you take this medicine for a long time then you \underline{would probably} develop a lot of other physical problems. \underline{Which} of these did you experience ?''}\\
This post 
evokes a lot of uncertainty, 
and does not specifically point to the occurrence of any side effect from a first-hand experience. Note the usage of strong modals (depicting a high degree of uncertainty)
``can'', ``may'', ``would'',
indefinite determiner ``some'',  
conditional ``if'', adverbs of possibility ``probably'' and the question particle ``which''. Even the usage of too many named entities, as in drug and disease names, in the post (refer the introductory example) can be detrimental for statement credibility.

Contrast the above post with the following one :\\
{\small ``Depo is very dangerous as a birth control and has too many long term side-effects like reducing bone density. \underline{Hence}, I \underline{will} never recommend anyone using \underline{this} as a birth control. \underline{Some} women tolerate it well but \underline{those} are \underline{the} minority. \underline{Most} women have horrible long lasting side-effects from it.''}\\
This post uses inferential conjunctions like ``hence'' drawing conclusions from an argument in the previous text segment, uses definite determiners ``this'', ``those'', ``the'' and ``most'' to pinpoint entities and the 
 highly certain weak
  modal ``will''.

Table~\ref{tab:linguistic} shows a set of linguistic features that are suitable
to discriminate between these two kinds of posts. Many of the features related to epistemic modality have been discussed in prior works in linguistics (\cite{Coates1987, Westney1986}). The feature types and values related to discourse coherence have been used in our earlier work (\cite{Mukherjee2012}), as well as \cite{Wolf2004}.

For each stylistic feature type $f_i$ shown in Table \ref{tab:linguistic} and each post $p_j$,
we compute the relative frequency of words of type $f_i$ occurring in $p_j$, thus
constructing a feature vector $F^L(p_j) =\langle freq_{ij} = \#(words~in~f_i) ~ / ~length(p_j) \rangle$.
We further aggregate these vectors over all posts $p_j$ by a user $u_k$ into
\begin{equation}
F^L(u_k) = \langle {\sum_{p_j~by~u_k} \#(words~in~f_i)}
~ / ~ {\sum_{p_j~by~u_k} length(p_j)} \rangle
 \label{eq2}
\end{equation}

\begin{table}
\centering
\small
\begin{tabular}{p{2.5cm}p{4cm}c}
\toprule
{\bf Feature types} & {\bf Example values}\\\midrule
Strong modals & might, could, can, would, may\\\sectionrule
Weak modals & should, ought, need, shall, will\\\sectionrule
Conditionals & if\\\sectionrule
Negation & no, not, neither, nor, never\\\sectionrule
Inferential conj. & therefore, thus, furthermore\\\sectionrule
Contrasting conj. &  until, despite, in spite, though\\\sectionrule
Following conj. & but, however, otherwise, yet\\\sectionrule
Definite det. & the, this, that, those, these\\\sectionrule
First person & I, we, me, my, mine, us, our\\\sectionrule
Second person & you, your, yours\\\sectionrule
Third person & he, she, him, her, his, it, its\\\sectionrule
Question particles & why, what, when, which, who\\\sectionrule
Adjectives & correct, extreme, long, visible\\\sectionrule
Adverbs &maybe, about, probably, much\\\sectionrule
Proper nouns & Xanax, Zoloft, Depo-Provera\\
\bottomrule
\end{tabular}
\caption{Stylistic features.}
\label{tab:linguistic}
\end{table}

\xhdrNoPeriod{Affective Features}
Each user has an \textit{affective state} that depicts her attitude and emotions that are reflected in 
her posts. Note that a user's affective state may change over time; so it is a property of
posts, not of users per se.
As an example, consider the following post:
{\small ``I've had chronic \underline{depression} off and on since adolescence. In the past I've taken Paxil (made me \underline{anxious}) and Zoloft (caused insomnia and stomach problems, but at least I was mellow ). I have been taking St. John's Wort for a few months now, and it helps, but not enough. I wake up almost every morning feeling very \underline{sad} and \underline{hopeless}. As afternoon approaches I start to feel better, but there's almost always at least a low level of \underline{depression} throughout the day. ''}\\
The high level of depression and negativity in the post makes one wonder if the statements on 
drug side-effects are really credible. Contrast this post to the following one:\\
{\small ``A diagnosis of GAD (Generalized Anxiety Disorder) is made if you suffer from excessive anxiety or worry and have at least three symptoms including ... If the symptoms above, touch a chord with you, do speak to your GP. There are effective treatments for GAD, and Cognitive Behavioural Therapy in particular can help you ...''}
where the user objectivity and positivity in the post make it much more credible.

We use the WordNet-Affect 
lexicon~\cite{WNAffect}, where each word sense (WordNet synset) is mapped to one of 285 attributes
of the affective feature space, like \textit{confusion, ambiguity, hope, anticipation, hate, etc.}. 
We do not perform word sense disambiguation (WSD), and instead simply take
the most common sense of a word (which is generally a good heuristics for WSD). 
For each post, we create an affective feature vector $\langle F^E(p_j) \rangle$ using these features,
analogous to the stylistic vectors $\langle F^L(p_j) \rangle$. Table~\ref{tab:aff} shows a snapshot of the affective features used in this work.

\begin{table}
\centering
\small
\begin{tabular}{p{8.5cm}}
\toprule
{\bf Sample Affective Features}\\\midrule
affection, antipathy, anxiousness, approval, compunction, confidence, contentment, coolness, creeps, depression, devotion, distress, downheartedness, eagerness, edginess, embarrassment, encouragement, favor, fit, fondness, guilt, harassment, humility, hysteria, ingratitude, insecurity, jitteriness, levity, levitygaiety, malice, misery, resignation, selfesteem, stupefaction, surprise, sympathy, togetherness, triumph, weight, wonder\\
\bottomrule
\end{tabular}
\caption{Affective features.}
\label{tab:aff}
\end{table}

\newcommand{\helpfulness}{helpfulness\xspace}
\newcommand{\helpful}{helpful\xspace}

\xhdrNoPeriod{Preliminary Feature Exploration}
To test whether the linguistic features introduced so far are sufficiently informative of how \helpful a user is in the context of health forums, we conduct a preliminary experimental study.
In the {\small\tt \href{http://www.healthboards.com/}{healthboards.com}} forum,  community members have the option of
expressing their gratitude to a user if they find one of her posts helpful by giving ``thanks'' votes.
We use the {\em total number of ``thanks'' votes} that a user received from all her posts
as a weak proxy measure for user \helpfulness solely for the purpose of this preliminary experiment.

We train a regression model on the per-user stylistic feature vectors $F^L(u_k)$ with
\#thanks normalized by \#posts for each user $u_k$ as response variable. We repeat the same experiment using only the per-user affective feature vectors $F^E(u_k)$ to identify the most important affective features.

Figure~\ref{fig:langFeat} shows the relative weight of various stylistic and affective linguistic features in determining user \textit{helpfulness}, with positive weights being indicative of features contributing to a user receiving thanks in the community.
Figure~\ref{fig:affFeat} shows that user confidence, pride, affection and positivity in statements are correlated with user \helpfulness, in contrast to misery, depression and negativity in attitude. Figure~\ref{fig:styFeat} shows that inferential statements about definite entities have a positive impact, as opposed to the use of hypothetical statements, contrasting sentences, doubts and queries.

This experiment confirms that linguistic features can be informative of how \helpful a user is.  Although we use ``thanks'' votes as a proxy for user \helpfulness, there is no guarantee that the information provided by \helpful users is actually correct.  A user can get ``thanks'' for a multitude of reasons (e.g. being compassionate, supportive etc.), and yet provide incorrect information. Hence, while the features described here are part of our final model, the feature weights learned in this preliminary experiment will not be used; instead, partially provided expert information is used to train our probabilistic model (refer Section~\ref{sec:inference} for details).

\begin{figure*}
\begin{subfigure}{.6\textwidth}
\includegraphics[scale=0.42]{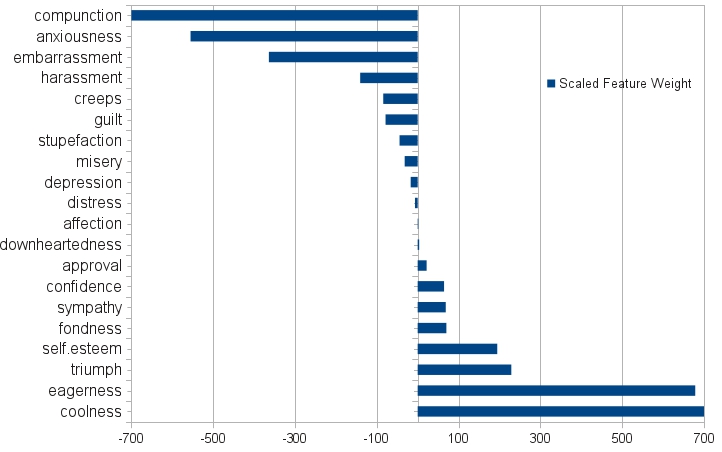}
\caption{Weight of top 20 affective features.}
\label{fig:affFeat}
\end{subfigure}
\begin{subfigure}{.4\textwidth}
\includegraphics[scale=0.55]{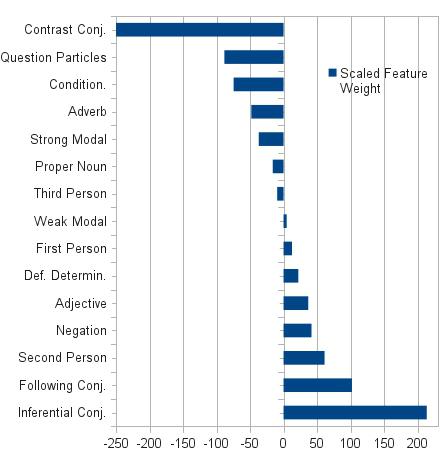}
\caption{Weight of stylistic features.}
\label{fig:styFeat}
\end{subfigure}
\caption{Relative importance of linguistic features for predicting user \helpfulness.}
\label{fig:langFeat}
\vspace*{-4mm}
\end{figure*}

\comment{
\begin{table}
\centering
\small
\begin{tabular}{lc}
\toprule
{\bf User Feature} & {\bf Scaled Feature Weight}\\\midrule
Replies / Queries & 305.72 \\
Gender (Female) & 159.43 \\
Posts & 137.09 \\
Thanks & 123.62 \\\bottomrule
\end{tabular}
\caption{User feature importance for predicting  \helpfulness.}
\label{tab:useFeat}
\vspace*{-4mm}
\end{table}
}


\subsection{User Features}

User demographics like age, gender, location, etc. and engagement in the community 
like number of posts, questions, replies, or thanks received,
 are expected to correlate with user authority in social networks.
Also, users who write long posts tend to deviate from the topic, often
with highly emotional digression. On the other hand, short posts can be regarded as being crisp, objective and
on topic.
We attempt to capture these intuitive aspects as additional 
per-user features $<F^U(u_k)>$. Specifically, we compute the first three moments of each user's
post-length distribution, in terms of sentences and in terms of words.

\section{Probabilistic Inference}
\label{sec:inference}

As outlined in Section~\ref{sec:model}, we model our learning task as a Markov Random Field (MRF), where the random variables are the users $U = \{u_1,u_2, ... u_{|U|}\}$, their posts $P = \{p_1, p_2 ... p_{|P|}\}$,
and the distinct statements $S = \{s_1, s_2 ... s_{|S|}\}$ 
about drug side-effects derived from all posts.
Our model is semi-supervised in that we harness ground-truth labels for a subset of statements, derived from the expert databases.
Let $S^L$ be the set of statements labeled by an expert as true or false, and let $S^U$ be the set of unlabeled statements. Our goal is to infer labels for the statements in $S^U$.  

The cliques in our MRF are 
triangles consisting of a statement $s_i$, a post $p_j$ that contains the statement, and a user $u_k$ who made this post.
As the same statement can be made in different posts by the same or other users, there are more cliques than statements. For convenient notation, let $S^*$ denote the set of statement instances that correspond to the set of cliques (with statements ``repeated'' when necessary).

Let $\phi_i(S^*_i, p_j, u_k)$ be a potential function for clique $i$. 
Each clique has a set of associated feature functions $F_i$ with a weight vector $W$. We denote the individual features and their weights as $f_{il}$ and $w_{l}$. The features are constituted by the stylistic, affective, and user features explained in Section~\ref{sec:features}:
$  F_i = F^L(p_j) \cup F^E(p_j) \cup  F^U(u_k)  $

Instead of computing the joint probability distribution $Pr(S,P,U;W)$ like in a standard MRF, we adopt the paradigm of Conditional Random Fields (CRF's)
and settle for the simpler task of estimating the 
conditional distribution:

\begin{equation} \label{eq4}
 Pr( S | P, U; W) = \frac{1}{Z(P,U)} \prod_i\phi_i(S^*_i, p_j, u_k; W)
\end{equation}

\noindent with normalization constant $Z(P,U)$,
or with features and weights made explicit:

\begin{equation} \label{eq5}
 Pr( S | P, U; W) = \frac{1}{Z(P,U)} \prod_i \exp (\sum_l w_l \times f_{il}(S^*_i, p_j, u_k))
\end{equation}

CRF parameter learning usually works on fully observed training data.
However, in our setting, only a subset of the $S$ variables have labels and we need to consider the partitioning of $S$ into $S^L$ and $S^U$:

\begin{equation} \label{eq6}
 Pr( S^U, S^L | P, U; W) = \frac{1}{Z(P,U)} \prod_i \exp (\sum_l w_l \times f_{il}(S^*_i, p_j, u_k))
\end{equation}

For parameter estimation, we need to maximize the marginal log-likelihood:

\begin{equation} \label{eq7}
 LL(W) = \log Pr(S^L|P,U;W) = \log \sum_{S^U} Pr(S^L, S^U | P,U; W)
\end{equation}

We can clamp the values of $S^L$ to their observed values in the training data~\cite{Sutton2012, Zhu2005} and compute the distribution over $S^U$ as:

\begin{equation} \label{eq8}
\small
 Pr(S^U|S^L,P,U; W) = 
 \frac{1}{Z(S^L,P,U)} \prod_i \exp (\sum_l w_l \times f_{il}(S^*_i, p_j, u_k))
\end{equation}

There are different ways of addressing the optimization problem for finding the argmax of $LL(W)$. In this work, we choose the Expectation-Maximization (EM) approach~\cite{mccullum2005}.
We first estimate the labels of the variables $S^U$ from the posterior distribution using Gibbs sampling, and then maximize the log-likelihood to estimate the feature weights.

\label{eq10}
\begin{subequations}
\small
\begin{flalign} \label{eq10.1}
 E-Step: q(S^U) &= Pr(S^U|S^L,P,U;W^{(\nu)}) &
\end{flalign}
\begin{equation} \label{eq10.2}
 M-Step: W^{(\nu+1)} = argmax_{W^{\prime}} \sum_{S^{U}} q(S^{U})\log Pr(S^L, S^{U}|P,U;W^{\prime})
\end{equation}
\end{subequations}

The update step to sample the labels of $S^U$ variables by Gibbs sampling is given by :

\begin{equation} \label{eq11}
 Pr(S^U_i| P,U,S^L; W) \propto \prod_{\nu\in C} \phi_\nu (S^*_\nu, p_j, u_k; W)
\end{equation}

\noindent where 
$C$ denotes the set of cliques containing statement $S^U_i$.

For the M-step in Equation~\ref{eq10.2}, we use an {$L_2$-regularized} Trust Region Newton Method~\cite{Lin2008}, suited for large-scale unconstrained optimization, where many feature values may be zero. For this we use an implementation of LibLinear~\cite{LibLinear}.

The above approach captures user trustworthiness implicitly via the weights of the feature vectors. However, we may want to model user trustworthiness in a way that explicitly aggregates over all the statements made by a user.
Let $t_k$ denote the trustworthiness of user $u_k$, measured as the fraction of statements made by him that were considered true
in the previous EM iteration. Let $S_{i,k}$ be the label assigned to $S_i$ by $u_k$, out of the statements $S_k$ made by him.

\begin{equation} \label{eq12}
  t_k = \frac{\sum_i \mathbbm{1}_{S_{i,k}= +1}}{|S_k|}
\end{equation}

Equation~\ref{eq11} can then be modified as:

\begin{equation} \label{eq13}
 Pr(S^U_i| P,U,S^L; W) \propto \prod_{\nu\in C} t_k \times \phi_\nu (S^*_\nu, p_j, u_k; W)
\end{equation}

Therefore, the random variable for trustworthiness depends on the proportion of \textit{true} statements made by the user. The \textit{label} of a statement, in turn, is determined by the language objectivity of the posts and trustworthiness of all the users in the community that make the statement.

The inference is an iterative process consisting of the following $3$ main steps :

\begin{enumerate}
 \item Estimate user trustworthiness $t_k$ using Equation~\ref{eq12}.
 \item Apply the {\emph E-Step} to estimate  $q(S^U;W^{(\nu)})$ \\
       For each $i$, sample $S_i^U$ from Equation~\ref{eq10.1} and~\ref{eq13}.
 \item Apply the {\emph M-Step} to estimate $W^{(\nu+1)}$ using Equation~\ref{eq10.2}.
\end{enumerate}

\section{Experimental Evaluation}
\label{sec:experiments}

In this section, we study the predictive power of our probabilistic
model and compare it to three baselines.

\subsection{Data}

We use data from the {\small\tt \href{http://www.healthboards.com}{healthboards.com}} community.
This is one of the largest online communities about health,
with 
$850,000$ registered members
and over $4.5$ million posted messages. 
We extracted $15,000$ users and all of their posts, $2.8$\ million posts in total.
Users are sampled based on their post frequency. Table~\ref{tab:author} shows the user categorization in terms of their community engagement. 
We employ a tool developed by~\cite{Ernst2014} to extract side-effect statements from the posts.
Details of the experimental study are available on our website.\footnote{\small \tt \href{http://www.mpi-inf.mpg.de/impact/peopleondrugs/}{http://www.mpi-inf.mpg.de/impact/peopleondrugs/}}

\begin{table}
\centering
{\small
\begin{tabular}{p{2.2cm}p{1.2cm}p{1.3cm}p{1.2cm}p{1.2cm}}
\toprule
{\bf Member Type}&{\bf Members}&{\bf Posts}&{\bf Average Qs.}&{\bf Average Replies}\\\midrule
Administrator & 1 & - & 363 & 934\\
Moderator & 4 & - & 76 & 1276\\
Facilitator & 16 & > 4700 & 83 & 2339\\
Senior Veteran & 966 & > 500 & 68 & 571\\
Veteran & 916 & > 300 & 41 & 176\\
Senior Member & 4321 & > 100 & 24 & 71\\
Member & 5846 & > 50 & 13 & 28\\
Junior Member & 1423 & > 40 & 9 & 18\\
Inactive & 1433 & - & -& -\\
Registered User & 70 & - & - & -\\
\bottomrule
\end{tabular}
}
\caption{User statistics.}
\label{tab:author}
\vspace*{-4mm}
\end{table}

As ground truth for drug side-effects, we rely on data from the Mayo Clinic portal.\footnote{\small\tt \href{http://www.mayoclinic.org/drugs-supplements/}{mayoclinic.org/drugs-supplements/}}
The website contains curated expert information about drugs, with
side-effects being listed as \emph{more common, less common and rare}
for each drug.
We extracted $2,172$\ drugs which are categorized into $837$ drug\ families.
For our experiments, we select $6$ widely used drug families,
based on {\small\tt \href{http://www.webmd.com}{webmd.com}}
Table~\ref{tab:drug} gives information on this sample
and its coverage in the posts at {\small\tt \href{http://www.healthboards.com/}{healthboards.com}}. Table~\ref{tab:SE} shows the number of common, less common, and rare side-effects
for the six drug families as given by the Mayo Clinic portal.


\subsection{Baselines}

We compare our probabilistic model against the following baseline methods,
using the same features that our model has available. Each model ranks the same set of side-effect candidates from the IE engine~\cite{Ernst2014}.

\begin{table}
\centering
{\small
\begin{tabular}{p{2.0cm}p{3.8cm}p{0.7cm}p{0.7cm}}
\toprule
{\bf Drugs}&{\bf Description}&{\bf Users}&{\bf Posts}\\\midrule
alprazolam, niravam, xanax & relieve symptoms of anxiety, depression, panic disorder & 2785 & 21,112 \\\midrule
ibuprofen, advil, genpril, motrin, midol, nuprin & relieve pain, symptoms of arthritis, such as inflammation, swelling, stiffness, joint pain & 5657 & 15,573\\\midrule
omeprazole, prilosec &treat acidity in stomach, gastric and duodenal ulcers, \dots & 1061 & 3884\\\midrule
metformin, glucophage, glumetza, sulfonylurea &treat high blood sugar levels, sugar diabetes & 779 & 3562\\\midrule
levothyroxine, tirosint &treat hypothyroidism: insufficient hormone production by thyroid gland & 432 & 2393 \\\midrule
metronidazole, flagyl &treat bacterial infections in different body parts & 492 & 1559\\
\bottomrule
\end{tabular}
}
\caption{Information on sample drug families.}
\label{tab:drug}
\vspace*{-2mm}
\end{table}

\begin{table}
\centering
\small
\begin{tabular}{lccc}
\toprule
{\bf Drug family}&{\bf Common}&{\bf Less common}&{\bf Rare}\\\midrule
alprazolam & 35 & 91 & 45\\
ibuprofen & 30 & 1 & 94\\
omeprazole & - & 15 & 20\\
metformin & 24 & 37 & 5\\
levothyroxine & - & 51 & 7\\
metronidazole & 35 & 25 & 14 \\
\bottomrule
\end{tabular}
\caption{Number of common, less common, and rare side-effects listed by experts.}
\label{tab:SE}
\vspace*{-4mm}
\end{table}

\xhdrNoPeriod{Frequency Baseline}
For each statement on a drug side-effect, we consider how frequently the statement
has been made in community. This gives us a ranking of side-effects.

\xhdrNoPeriod{SVM Baseline}
For each drug and each of its possible side-effect, we determine all posts where it
is mentioned and aggregate the features $F^L$, $F^E$, $F^U$,
described in  Section~\ref{sec:features}, over all these posts, to create a single feature vector for each side-effect.

We use the ground-truth labels from the Mayo Clinic portal to
train a Support Vector Machine (SVM) classifier with a linear kernel, $L_2$ Loss,
and $L_1$ or $L_2$ regularization, for classifying unlabeled statements.

\xhdrNoPeriod{SVM Baseline with Distant Supervision}
As the number of common side-effects for any drug is typically small, the above approach to create a single feature vector for each side-effect results in a very small training set. Hence, we use the notion of \textit{distant supervision} to create a rich, expanded training set.

A feature vector is created for \textit{every mention} or instance of a side-effect in different user posts. The feature vector
$<S_i, p_j, u_k>$ has the label of the side-effect, and represents the set of cliques in Equation~\ref{eq4}. The semi-supervised CRF formulation in our approach further allows for information sharing between the cliques to estimate the labels of the unobserved statements from the expert-provided ones.  This creates a noisy training set, as a post may contain multiple side-effects, positive and negative. This results in multiple similar feature vectors with different labels.

During testing, the same side-effect may get different labels from its different instances. We take a majority voting of the labels obtained by a side-effect, across predictions over its different instances, and assign a unique label to it.


\subsection{Experiments and Quality Measures}

We conduct two lines of experiments, with different settings on what is
considered ground-truth.

\xhdrNoPeriod{Experimental Setting I} 
We consider
only {\em most common side-effects} listed by the Mayo Clinic portal
as positive ground-truth, whereas all other side-effects (less common, rare and unobserved) are considered to be negative instances (i.e., so unlikely that they should be
considered as false statements, if reported by a user).
The training set is constructed in the same way.
This setting aims to study the predictive power of our model in determining the common side-effects of a drug,
in comparison to the baselines.


\xhdrNoPeriod{Experimental Setting II}
Here we address our original motivation: discovering less common and rare side-effects.
As positive ground-truth, we consider common and less common side-effects, as stated
by the experts on the Mayo Clinic site, whereas all rare and unobserved
side-effects are considered negative instances.
The model is trained with the same partitioning of true vs. false statements.
Our goal here is to test how well the model can identify \textit{less known} and \textit{rare}
side-effects as true statements. For the rare side-effects, we measure only the recall for such statements being labeled as true statements (if indeed deemed credible according to posts and users), although the training consider \textit{only} common and less common
side-effects as positive instances. Such rare statements are not incorporated as positive training examples, as users frequently talk about experiencing them in the community. Instead, the classifier is made to learn only from the most probable ones as positive instances.

\xhdrNoPeriod{Train-Test Data Split}\label{lab:ds}
For each drug family, we create multiple random splits of $80\%$ training data
and $20\%$ test data.  All results reported below
are averaged over $200$ such splits. 
All baselines and our CRF model use same test sets.

\xhdrNoPeriod{Evaluation Metrics} The standard measure for the quality of a binary classifier is {\em accuracy}:
$\frac{tp+tn}{tp+fn+tn+fp}$.
We also report the \textit{specificity} ($\frac{tn}{tn+fp}$) and \textit{sensitivity} ($\frac{tp}{tp+fn}$).
Sensitivity measures the true positive rate or the model's ability to identify positive side-effects, whereas specificity measures true negative rate.

\subsection{Results and Discussions}

Table~\ref{tab:Acc} shows the accuracy comparison of CRF with the baselines for different drug families in the first setting. The first naive baseline, which simply considers the frequency of posts containing the side-effect by different users, has an average accuracy of $57.65\%$ across different drug families.

Incorporating supervision in the classifier as the first SVM baseline, along with a rich set of features for users, posts and language, achieves an average accuracy improvement of $11.4\%$. In the second SVM baseline, we represent each post of a side-effect by a user (<$S_i, p_j, u_k$>) as a separate feature vector. This not only expands the training set leading to better parameter estimation, but also represents the set of cliques in Equation~\ref{eq4}. This has an average accuracy improvement of $7\%$ for an SVM with $L_1$ regularization and $9\%$ for corresponding $L_2$ regularization. The CRF model, considering the coupling between users, posts and statements, allows information to flow between the cliques in a feedback loop giving a further accuracy improvement of $6\%$ over the {\emph SVM $L_1$} baseline and $4\%$ over the $L_2$ baseline.

\begin{table}
\centering
\small
\begin{tabular}{lccccc}
\toprule
\multicolumn{ 1}{l}{\bf Drugs} & \multicolumn{ 1}{c}{\bf \parbox{1cm}{Post\\Freq.}} & \multicolumn{ 3}{c}{\bf SVM} & \multicolumn{ 1}{c}{\bf CRF} \\ \cmidrule{ 3- 5}
\multicolumn{ 1}{l}{} & \multicolumn{ 1}{c}{} & \multicolumn{ 1}{l}{\bf w/o DS} & \multicolumn{ 2}{c}{\bf DS} & \multicolumn{ 1}{l}{} \\ \cmidrule{ 4- 5}
\multicolumn{ 1}{l}{} & \multicolumn{ 1}{c}{} & \multicolumn{ 1}{l}{} & $L_1$ & $L_2$ & \multicolumn{ 1}{l}{} \\ \midrule
Alprazolam & 57.82 & 70.24 & 73.32 & 73.05 & 79.44  \\
Metronidazole & 55.83 & 68.83 & 79.82 & 78.53 & 82.59  \\
Omeprazole & 60.62 & 71.10 & 76.75 & 79.15 & 83.23  \\
Levothyroxine & 57.54 & 76.76 & 68.98 & 76.31 & 80.49 \\
Metformin & 55.69 & 53.17 & 79.32 & 81.60 & 84.71 \\
Ibuprofen & 58.39 & 74.19 & 77.79 & 80.25 & 82.82 \\
\bottomrule
\end{tabular}
\caption{Accuracy comparison in setting I.}
\label{tab:Acc}
\vspace*{-2mm}
\end{table}

\begin{table}
\centering
\small
\begin{tabular}{p{1.7cm}p{1.3cm}p{1.3cm}p{1.3cm}p{0.8cm}}
\toprule
{\bf Drugs}&{\bf \parbox{1.5cm}{Sensi-\\tivity}}&{\bf \parbox{1.5cm}{Speci-\\ficity}}&{\bf \parbox{1.3cm}{Rare SE\\Recall}}&{\bf \parbox{0.8cm}{Acc-\\uracy}} \\\midrule
Metformin & 79.82 & 91.17 & 99 & 86.08\\
Levothyroxine & 89.52 & 74.5 & 98.50 & 83.43\\
Omeprazole & 80.76 & 88.8 & 89.50 & 85.93\\
Metronidazole & 75.07 & 93.8 & 71 & 84.15\\
Ibuprofen & 76.55 & 83.10 & 69.89 & 80.86\\
Alprazolam & 94.28 & 68.75 & 61.33 & 74.69\\
\bottomrule
\end{tabular}
\caption{CRF performance in setting II.}
\vspace*{-4mm}
\label{tab:rare}
\end{table}

Figure~\ref{fig:modelComparison} shows the sensitivity and specificity comparison of the baselines with the CRF model. Our approach has an overall $5\%$ increase in sensitivity and $3\%$ increase in specificity over the {\emph SVM $L_2$} baseline.

The specificity increase over the {\emph SVM $L_2$} baseline is maximum for the Alprazolam drug family at $8.33\%$ followed by Levothyroxine at $4.6\%$. The users taking anti-depressants like Alprazolam suffer from anxiety disorder, panic attacks, depression etc. and report a large number of side-effects of drugs. Hence, it is very difficult to negate certain side-effects, in which our model performs very well due to well-designed language features. Also, Alprazolam and Levothyroxine have a large number of expert-reported side-effects (refer Table~\ref{tab:SE}) and corresponding user-reported ones, and the model performs a good learning for the negative class.

The drugs Metronidazole, Metformin and Omeprazole treat some serious physical conditions, have less number of expert and user-reported side-effects. Consequently, our model captures user statement corroboration well to attain a sensitivity improvement of $7.89\%, 6.5\%$ and $6.33\%$ respectively. Overall, classifier performs the best in these drug categories.

Table~\ref{tab:rare} shows the overall model performance, as well as the recall for identifying rare side-effects of each drug in the second setting.
The drugs Metformin, Levothyroxine and Omeprazole have much less number of side-effects, and the classifier does an almost perfect job in identifying all of them. Overall, the classifier has an accuracy improvement of $2-3\%$ over these drugs in Setting II. However, the classifier accuracy {\emph significantly} drops for the anti-depressants (Alprazolam) after the introduction of ``less common'' side-effects as positive statements in Setting II. The performance drop is attributed to the loss of $8.42\%$ in specificity due to increase in the number of false-positives, as there is conflict between what the model learns from the language features (about negative side-effects) and that introduced as ground-truth.

\xhdrNoPeriod{Feature Informativeness} In order to find the \textit{predictive power} of individual feature classes, tests are perfomed using $L_2$-loss and $L_2$-regularized Support Vector Machines over a split of the test data. Affective features are found to be the most informative, followed by document length statistics, which are more informative than user and stylistic features. Importance of document length distribution strengthens our observation that objective posts tend to be crisp, whereas longer ones often indulge in emotional digression.

Amongst the user features, the most significant one is the ratio of the number of replies by a user to the questions posted by her in the community, followed by the gender, number of posts by the user and finally the number of thanks received by her from fellow users. There is a gender-bias in the community, as $77.69\%$ active contributors in this health forum are female.

Individual F-scores of the above feature sets vary from $51\%$ to $55\%$ for Alprazolam; whereas the combination of all features yield $70\%$ F-score.

\begin{figure}
\includegraphics[scale=0.27]{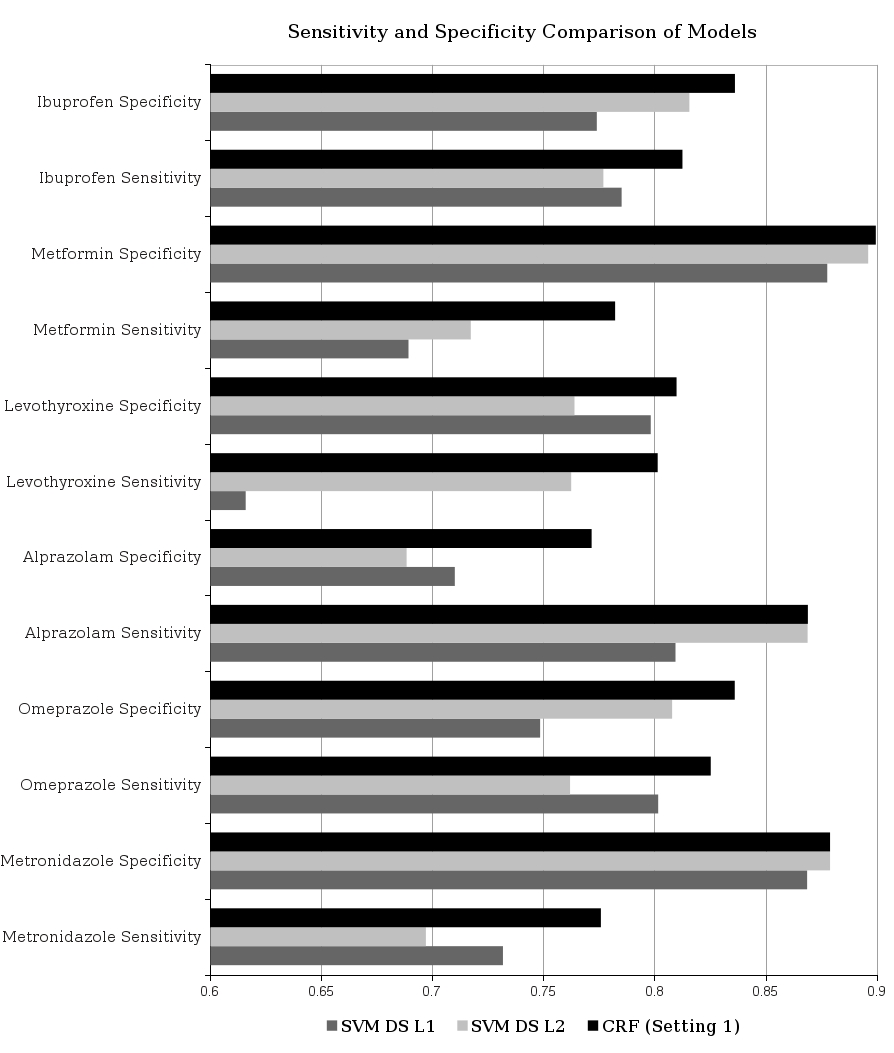}
\caption{Specificity and sensitivity comparison of models.}
\label{fig:modelComparison}
\vspace*{-4mm}
\end{figure}

\section{Use-Case Experiments}
\label{sec:usecases}

Section~\ref{sec:experiments} has focused on evaluating the predictive power of our model and inference method. Now we shift the focus to two application-oriented use-cases: 1) discovering side-effects of drugs that are not covered by expert databases, and 2) identifying the most trustworthy users that one would want to follow for certain topics.

\subsection{Discovering Rare Side Effects}
\label{subsec:usecaseRare}

Members of an online community may report side-effects that are either flagged as very rare in an expert knowledge base (KB) or not listed at all. We call the latter {\em out-of-KB} statements. As before, we use the data from {\small\tt mayoclinic.org} as our KB, and focus on the following 2 drugs representing different kinds of medical conditions and patient-reporting styles: Alprazolam and Levothyroxine. For each of these drugs, we perform an experiment as follows.

For each drug $X$, we use our IE machinery to identify all side-effects $S$ that are reported for $X$, regardless of whether they are listed for $X$ in the KB or not. The IE method uses the set of all side-effects listed for {\em any} drug in the KB as potential result. For example, if ``hallucination'' is listed for some drug but not for the drug Xanax, we capture mentions of hallucination in posts about Xanax.
We use our probabilistic model to compute credibility scores for these out-of-KB side-effects, and compile a ranked list of 10 highest-scoring side-effects for each drug. This ranked list is further extended by 10 randomly chosen out-of-KB side-effects (if reported at least once for the given drug).

The ranked list of out-of-KB side-effects is shown to $2$ expert annotators who manually assess their credibility, by reading the complete discussion thread (e.g. expert replies to patient posts) and other threads that involve the users who reported the side-effect. The assessment is binary: true (1) or false (0); we choose the final label as majority of judges. This way, we can compute the quality of the ranked list in terms of the {\em NDCG (Normalized Discounted Cumulative Gain~\cite{kalervo2002})} measure
$NDCG_p = \frac{DCG_p}{IDCG_p}$ where
\begin{equation}
DCG_p = rel_1 + \sum_{i=2}^p \frac{rel_i}{\log_2 i}
\end{equation}

Here, $rel_i$ is the graded relevance of a result ($0$ or $1$ in our case) at position $i$. DCG penalizes relevant items appearing lower in the rank list, where the graded relevance score is reduced logarithmically proportional to the position of the result. As the length of lists may vary for different queries, DCG scores are normalized using the ideal score, IDCG
where the results of a rank list are sorted by relevance giving the maximum possible DCG score.
We also report the inter-annotator agreement using Cohen's Kappa measure.

Table~\ref{tab:useRare} shows the Kappa and NDCG score comparison between the baseline and our CRF model.
The baseline here is to rank side-effects by frequency \textit{i.e.}
how often are they reported in the posts of different users on the given drug.
The strength of Kappa is considered ``moderate'' (but significant), which depicts the difficulty in identifying the side-effects of a drug just by looking at user posts in a community. The baseline performs very poorly for the anti-depressant Alprazolam, as the users suffering from anxiety disorders report  a large number of side-effects most of which are not credible. On the other hand, for Levothyroxine (a drug for hypothyroidism), the baseline model performs quite well as the users report more serious symptoms and conditions associated with the drug, which also has much less expert-stated side-effects compared to Alprazolam (refer Table~\ref{tab:drug}). The CRF model performs perfectly for both drugs.

\begin{table}
\centering
\small
\begin{tabular}{cccc}
\toprule
\multicolumn{ 1}{c}{Drug} & \multicolumn{ 1}{c}{Kappa} & \multicolumn{ 2}{c}{Model NDCG Scores} \\ \cmidrule{ 3- 4}
\multicolumn{ 1}{c}{} & \multicolumn{ 1}{c}{} & Frequency & CRF \\
Alprazolam, Xanax & \multicolumn{1}{c}{0.471} & \multicolumn{1}{c}{0.31} & \multicolumn{1}{c}{1} \\
Levothyroxine, Tirosint & \multicolumn{1}{c}{0.409} & \multicolumn{1}{c}{0.94} & \multicolumn{1}{c}{1} \\
\bottomrule
\end{tabular}
\caption{\small Use-Case experiment on rare drug side-effects}
\label{tab:useRare}
\end{table}

\begin{table}
\centering
\small
\begin{tabular}{cccc}
\toprule
\multicolumn{ 1}{c}{Drug} & \multicolumn{ 1}{c}{Kappa} & \multicolumn{ 2}{c}{Model NDCG Scores} \\ \cmidrule{ 3- 4}
\multicolumn{ 1}{c}{} & \multicolumn{ 1}{c}{} & Frequency & CRF \\
Alprazolam, Xanax & \multicolumn{1}{c}{0.783} & \multicolumn{1}{c}{0.82} & \multicolumn{1}{c}{1} \\
Levothyroxine, Tirosint & \multicolumn{1}{c}{0.8} & \multicolumn{1}{c}{0.57} & \multicolumn{1}{c}{0.81} \\
\bottomrule
\end{tabular}
\caption{\small Use-Case experiment on following trustworthy users}
\label{tab:useFollower}
\vspace*{-4mm}
\end{table}

\subsection{Following Trustworthy Users}

In the second use-case experiment, we evaluate how well our model can identify trustworthy users in a community. We find the top-ranked users in the community given by their trustworthiness scores ($t_k$), for each of the drugs Alprazolam and Levothyroxine. As a baseline model, we consider the top-thanked contributors in the community. The moderators and facilitators of the community, listed by both models as top users, are removed from the ranked lists, in order to focus on the interesting, not obvious cases.
Two judges are asked to annotate the top-ranked users listed by each model
 as trustworthy or not, based on the users' posts on the target drug. The judges are asked to mark a user trustworthy if they would consider following the user in the community. Although this exercise may seem highly subjective, the Cohen's Kappa scores
show high inter-annotator agreement. The strength of agreement is considered to be
``very good'' for the user posts on Levothyroxine, and ``good'' for the Alprazolam users.

The baseline model performs poorly for Levothyroxine.
The CRF model outperforms the baseline in both cases.

\section{Related Work}
\label{sec:relatedwork}

\noindent{\bf Information Extraction (IE):}
There is ample work on extracting Subject-Predicate-Object (SPO) like statements
from natural-language text. 
The survey \cite{Sarawagi2008} gives an overview; 
\cite{Krishnamurthy2009,Bohannon2012,SuchanekWeikum2013} provide
additional references. 
State-of-the-art methods combine pattern matching with 
extraction rules and consistency reasoning.
This can be done either in a shallow manner, over sequences
of text tokens, or in combination with deep parsing and
other linguistic analyses.
The resulting SPO triples often have highly varying confidence,
as to whether they are really expressed in the text or picked
up spuriously.
Judging the credibility of statements is out-of-scope for IE itself.

\xhdrNoPeriod{IE on Biomedical Text}
For extracting facts about diseases, symptoms, and drugs, 
customized IE techniques have been developed to tap biomedical
publications like PubMed articles.
Emphasis has been on the molecular level, i.e. proteins, genes,
and regulatory pathways 
(e.g., \cite{Bundschus2008,Krallinger2008,Bjoerne2010}), and
to a lesser extent on biological or medical events from scientific
articles and from clinical narratives~\cite{Jindal2013,Xu2012}.
\cite{Paul2013} has used LDA-style models for summarization
of drug-experience reports.
\cite{Ernst2014} has employed such techniques to build
a large knowledge base for life science and health. Recently, \cite{White2014} demonstrated how to derive insight on drug effects from query logs
of search engines. 
Social media has played a minor role in this prior
IE work.

\xhdrNoPeriod{Truth Finding}
This research direction aims to assess the truth of a given
statement that is frequently observed on the Web, a typical
example being ``Obama is a muslim''.
\cite{Yin2007,Pasternack2010,Pasternack2011} develop methods for statistical reasoning on the cues
for the statement being true vs. false.
\cite{Li2011} uses information-retrieval techniques to
systematically generate alternative hypotheses for a given
statement (e.g., ``Obama is a Christian''), and assess the
evidence for each alternative. \cite{Li2012} has developed
approaches for 
structured data such as flight times or stock quotes,
where different Web sources often yield contradictory values.
Recently, \cite{Pasternack2013} presented an LDA-style
latent-topic model for discriminating true from false claims,
with various ways of generating incorrect statements
(guesses, mistakes, lies).
\cite{KDD-DMH-2011} addressed  truth assessment for
medical claims about diseases and their treatments (including drugs and general phrases such as ``surgery''), by an IR-style evidence-aggregation and
ranking method
over \textit{curated} health portals.
Although these are elaborate models,
they are not geared for our setting where
the credibility of statements is intertwined
with the trust in users and the language of
user posts.
Moreover, none of these prior works have considered
online discussion forums.

\xhdrNoPeriod{Language Analysis for Social Media}
Work on sentiment analysis has looked into language features in customer reviews (e.g., \cite{Turney2002,PangLee2007,Liu2012,Mukherjee2012} and author writing style (\cite{Mukherjee2014}). Going beyond this special class of texts, \cite{GreeneResnik2009,Recasens2013} have studied the use of biased language in Wikipedia and similar collaborative communities. Even more broadly, the task of characterizing subjective language has been addressed, among others, in \cite{Wiebe2005,Lin2011}.
The work by \cite{Wiebe2011} has explored benefits between subjectivity analysis and information extraction. None of this prior work has addressed the specifics of discussions in online healthforums.

\xhdrNoPeriod{Trust and Reputation Management}
This area has received much attention, mostly motivated by
analyzing customer reviews for product recommendations, but also
in the context of social networks.
\cite{Kamvar2003,Guha2004} are seminal works that modeled
the propagation of trust within a network of users. 
TrustRank \cite{Kamvar2003} has become a popular measure
of trustworthiness, based on random walks on (or spectral
decomposition of) the user graph.
Reputation management has also been studied in the context
of peer-to-peer systems, the blogosphere, and online interactions
\cite{Adler2007,Agarwal2009,Despotovic2009,deAlfaro2011,Hang2013}.

All these works focused on explicit relationships
between users to infer authority and trust levels.
The only content-aware model for trust propagation,
and in fact closest to our work, is \cite{KDD2011}.
This work develops a HITS-style algorithm for propagating
trust scores in a heterogeneous network of claims, sources,
and documents.
Evidence for a claim is collected from related documents using generic IR-style word-level measures. 
Our work considers online users and rich language features for their posts. This more demanding setting
requires a more sophisticated model, like our CRF.

\section{Conclusion}

We propose a probabilistic graphical model to jointly learn the interactions between user trustworthiness, statement credibility and language use. We apply the model to extract side-effects of drugs from health communities, where we leverage the user interactions, stylistic and affective features of language use, and user properties to learn the credibility of user statements. We show that our approach is effective in reliably extracting side-effects of drugs and filtering out false information prevalent in online communities.

Our two-fold experiments, against expert knowledge on one hand, and
based on a user study on the other,
show that our model clearly outperforms all baselines.
The user study is designed to identify rare side-effects of drugs, a scenario where large-scale non-expert data has the potential to complement expert knowledge,
and to identify trustworthy users in the community one would want to follow for certain topics.

We believe our model can be a strong asset for possible in-depth analysis, like determining the specific conditions (age, gender, social group, life style, other medication, etc.) under which side-effects are observed.

Although our model achieves high accuracy in most of the test cases, 
it suffers from the usage of a simple Information Extraction (IE) machinery to identify possible side-effects in statements. The tool misses out on certain kinds of paraphrases (e.g. ``nightmares'' and ``unusual dream'' for Xanax) resulting in a drop in recall. We believe a more powerful IE approach can further boost the quality of our approach.

\comment{not referred in paper 8-21, 23,24,25,27-30,32,34-40}

\bibliographystyle{abbrev}

\end{document}